\documentclass[final]{cvpr}

\usepackage[pagebackref=true,breaklinks=true,colorlinks,bookmarks=false]{hyperref}

\usepackage{amsmath}
\usepackage{amsfonts}
\usepackage{bm}

\def\eqref#1{equation~\ref{#1}}
\def\1{\bm{1}}

\DeclareMathAlphabet{\mathsfit}{\encodingdefault}{\sfdefault}{m}{sl}
\SetMathAlphabet{\mathsfit}{bold}{\encodingdefault}{\sfdefault}{bx}{n}

\DeclareMathOperator*{\argmax}{arg\,max}

\newcommand{\parens}[1]{\left(#1\right)}
\newcommand{\brackets}[1]{\left[#1\right]}

\usepackage{times}
\usepackage[utf8]{inputenc} % allow utf-8 input
\usepackage[T1]{fontenc}    % use 8-bit T1 fonts
\usepackage{booktabs}       % professional-quality tableshttps://www.overleaf.com/project/5f10e663331ee100011f0852
\usepackage{nicefrac}       % compact symbols for 1/2, etc.
\usepackage{microtype}      % microtypography
\usepackage{caption}
\usepackage{subcaption}

\usepackage{epsfig}
\usepackage{amsmath}
\usepackage{amsfonts}
\usepackage{graphics}
\usepackage{graphicx}
\usepackage{color}
\usepackage{bbm}
\usepackage{graphicx}
\usepackage{wrapfig}
\usepackage{tabularx}
\usepackage{caption}
\usepackage[normalem]{ulem}
\usepackage{url}            % simple URL typesetting
\usepackage{multirow}
\usepackage{soul}
\usepackage{pifont}
\newcommand{\cmark}{\ding{51}}
\newcommand{\xmark}{\ding{55}}
\usepackage[title]{appendix}

\title{
Positive-Congruent Training: Towards Regression-Free Model Updates
}
\author{
Sijie Yan\thanks{Currently at The Chinese University of Hong Kong. Work conducted while at AWS.} \quad \ 
Yuanjun Xiong \quad \ 
Kaustav Kundu \quad \ 
Shuo Yang \quad \
Siqi Deng \quad \ \\
Meng Wang \quad \
Wei Xia \quad \
Stefano Soatto \\\\
AWS/Amazon AI\\
{\tt\small  yysijie@gmail.com}, \quad
{\tt\small \{yuanjx, kaustavk, shuoy, siqideng, mengw, wxia, soattos\}@amazon.com}
}
\date{Nov 17, 2020}

\begin{document}

\maketitle
\pagestyle{empty}
\thispagestyle{empty}

\begin{abstract}
Reducing inconsistencies in the behavior of different versions of an AI system can be as important in practice as reducing its overall error. 
In image classification, sample-wise inconsistencies appear as ``negative flips'': A new model incorrectly predicts the output for a test sample that was correctly classified by the old (reference) model. 
Positive-congruent (PC) training  aims at reducing error rate while at the same time reducing negative flips, thus maximizing congruency with the reference model only on positive predictions, unlike model distillation. We propose a simple approach for PC training, Focal Distillation, which enforces congruence with the reference model by giving more weights to samples that were correctly classified. We also found that, 
if the reference model itself can be chosen as an ensemble of multiple deep neural networks, negative flips can be further reduced without affecting the new model's accuracy. 
\end{abstract}

\section{Introduction}

\begin{figure}[th]
    \centering
    \includegraphics[width=.9\linewidth]{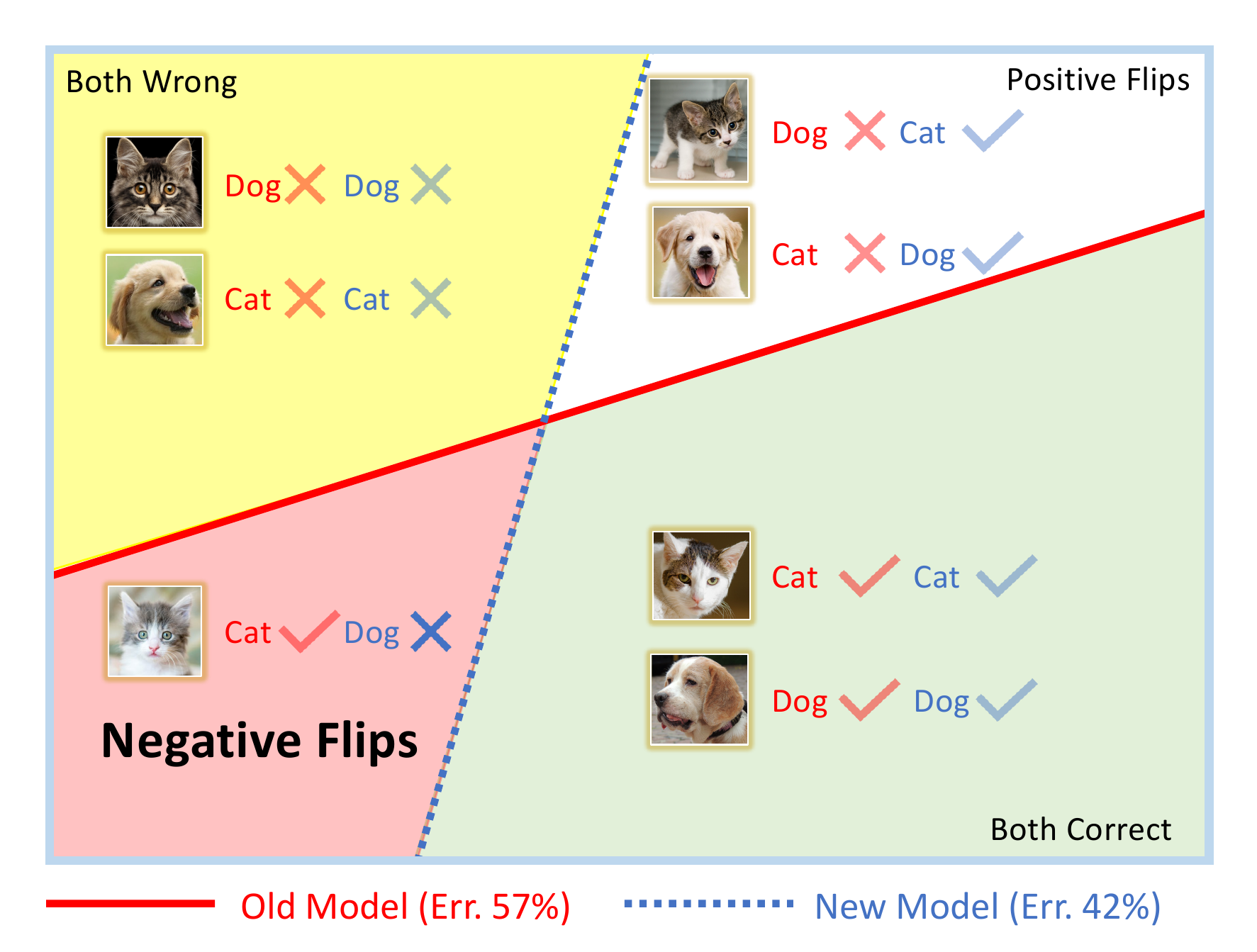}
    \caption{{\bf Regression in model update:} When updating an old classifier (red) to a new one (dashed blue line), we correct mistakes (top-right, white), but we also introduce errors that the old classifier did not make ({\bf negative flips}, bottom-left, red). While on average the errors decrease (from $57\%$ to $42\%$ in this toy example), regression can wreak havoc with downstream processing, nullifying the benefit of the update. }
    \label{fig:teaser}
\end{figure}

\begin{figure*}
    \centering
    \includegraphics[width=1.0\linewidth]{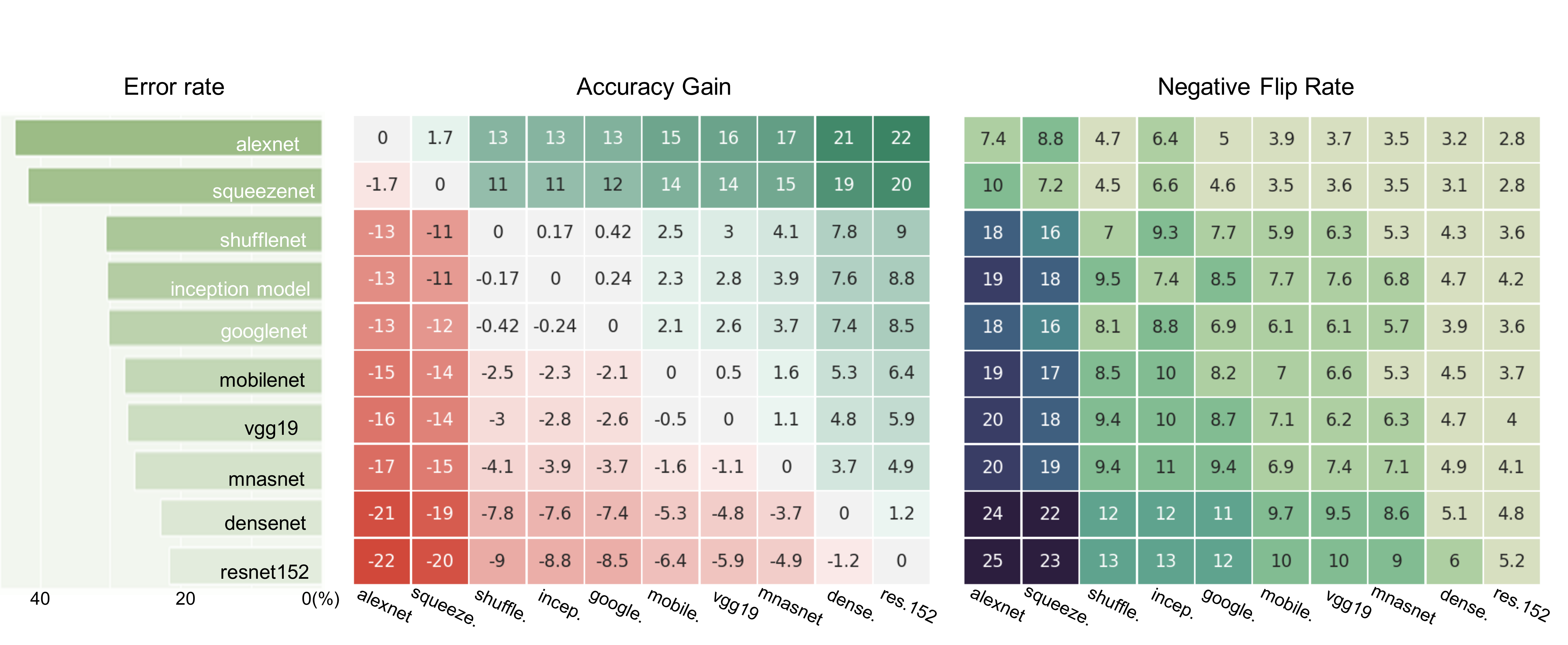}
    \caption{Differences in overall accuracy vs. negative flip rates. We measure the accuracy gains and negative flip rates on the ILSVRC12~\cite{ILSVRC15} validation set when we update from various old models ($y$-axis) to new models ($x$-axis). Multiple CNN architectures~\cite{krizhevsky2017imagenet, he2016resnet,zhang2020resnest} are used for comparison. First we observe training the same architecture twice can cause regression. If the new model has higher/lower capacity than the old one, the error rate decreases/increases (above/below the diagonal), but the NFR is always positive. Note that in some cases the NFR is of the same order of the error rate change.}
    \label{fig:acc_diff_vs_nfr}
\end{figure*}

Imagine a ``new and improved'' version of the software that manages your photo collection exhibiting mistakes absent in the old one. Even if the average number of errors has decreased, every new mistake on your old photos feels like a step backward (Fig.~\ref{fig:teaser}), leading to perceived {\em regression}.\footnote{
In the Software Industry, regression refers to the deterioration of performance after an update. Even if the updated model has on average better performance, any regression risks breaking post-processing pipelines, which is why trained models are updated sporadically despite the steady stream of improvements reported in the literature.} 
We tackle regression in the update of image classification models, where an old (reference) model is replaced by a new (updated) model. We call test samples that are correctly labeled by both the new and the old models {\em positive-congruent.}  On the other hand, {\em negative flips} are samples correctly classified by the old model but incorrectly by the new one. Their fraction of the total number is called negative flip rate (NFR), which measures regression. There are also {\em positive flips}, where the old model makes mistakes that the new one corrects, which are instead beneficial.  

Reducing the NFR could be accomplished by reducing the overall error rate (ER). However, reducing the ER is neither necessary nor sufficient to reduce the NFR. In fact, models trained on the same data with different initial conditions, data augmentations, and hyperparameters tend to yield similar error rates, but with {\em errors occurring on different samples.} For instance, two ResNet152 models trained on ImageNet with just different initializations achieve the same accuracy of $78.3\%$ but differ on $10.4\%$ of the samples in the validation set (Fig.~\ref{fig:acc_diff_vs_nfr}). Assuming an equal portion of positive and negative flips, updating from one model to another incurs a $5.2\% NFR$. This can be reduced simply by trading errors while maintaining an equal error rate. Thus, reducing the ER is not necessary to reduce the NFR. Reducing the ER to a value other than zero is not sufficient either: From  Fig.~\ref{fig:acc_diff_vs_nfr} we see that, when updating an old AlexNet~\cite{krizhevsky2017imagenet} to a new Resnet-152~\cite{he2016resnet}  ER reduces significantly, but we still suffer a $3\%$ NFR.

On the other hand, NFR can be made zero trivially by copying the old model, albeit with no ER reduction. Model distillation tries to bias the update towards the old model while reducing the ER. However, we wish to mimic the old model {\em only when it is right}. Thus, regression-free updates are not achieved by ordinary distillation.

Reducing the error rate and the NFR are two separate and independent goals. {\em We call any training procedure that aims to minimize both the error rate and the negative flip rate Positive-Congruent (PC) training,} or PCT for short.

We first propose a simple  method for PC training when the old model is given. In this case, PCT is achieved by minimizing an additional loss term along with the standard classification loss (empirical cross-entropy). We consider several variants for the additional loss, and find \emph{Focal Distillation} (FD), a variant of the distillation loss that we introduce to bias the model towards positive congruence (Sect. \ref{sec:distillation}), to be most effective (Sect. \ref{sec:experiments}).

We also explore the problem of  ``future-proofing'' the reference model to facilitate subsequent PC training. This  \emph{forward setting} pertains to selecting the reference model, rather than PC training new ones. Motivated by the observation above that different training instantiations can yield the same error rate with different erroneous samples, we propose  a simple approach using ensembles. We show that this results in lower regression. In practice, ensembles are not viable at inference time in large-scale applications due to the high cost. Nonetheless,  this approach can be used as a paragon for PC training of a single model in future works. 

Our {\bf contributions} can be summarized as follows: (i) We formalize the problem of quality regression in pairs of classifiers, and introduce the first method for positive congruent training (Sect. \ref{sec:formalization}); (ii) we propose a variant of  model distillation (Sect. \ref{sec:distillation}) to perform PC training of a deep neural network (DNN); (iii) we show that reference models can be adapted for future PC-Training by replacing a single model with an ensemble of DNNs (Sect. \ref{sec:ensembles}). We conduct experiments on large scale image classification benchmarks (Sect. \ref{sec:experiments}), providing both a baseline (Focal Distillation) and a paragon (Ensemble) for future evaluation.

\section{Related Work}
\label{sec:prior-work}

PC training relates to the general areas of {\bf continual learning} \cite{chen2018lifelong,kirkpatrick2017weightcons,silver2002task,  thrun1998lifelong}, incremental learning~\cite{prabhu2020greedy, li2017learning}, open set recognition~\cite{bendale2016towards, scheirer2012toward}, and sequential learning~\cite{goodfellow2013empirical, mccloskey1989catastrophic}. The goal is to evolve a model to incorporate additional data or concepts, as reviewed in~\cite{de2019continual, parisi2019continual}. Specifically, \cite{goodfellow2013empirical, kirkpatrick2017weightcons, lopez2017gradient, li2017learning, prabhu2020greedy, rebuffi2017icarl, zenke2017continual} aim at training with a growing number of samples, classes, and tasks while maintaining similar performance to previously learned classes/tasks~\cite{mccloskey1989catastrophic, ratcliff1990connectionist}.

PC training does not restrict the new model to reuse the old model weights or architecture. Instead, it allows changing them along with the training algorithm, loss functions, hyper-parameters, and training set. Unlike most work focused on reducing forgetting~\cite{chaudhry2018riemannian, li2017learning, lopez2017gradient}, PC training focuses on reducing NFR.

PC Training has strong connections to {\bf knowledge distillation}~\cite{hinton2015distilling} and weight consolidation~\cite{kirkpatrick2017weightcons}, that aim to keep a model close to a reference one, regardless of whether they are correct or not. PC training focuses on enforcing similarity only on cohorts that belong to negative flips. 

Along a similar vein, {\bf backward-compatible training} \cite{bct,srivastava2020an} aims to design new models that are inter-operable with old ones. PC training is a form of backward-compatibility, but prior literature still measures compatibility using error rate, rather than NFR. As shown in experiments, these methods do not directly help reduce the NFR. Implicit negative flips during training of a single model are studied in~\cite{toneva2018empirical}.  We also conduct experiments on the evolution of negative flips during training of the new model. 

\section{Negative Flips in Model Updates}
\label{sec:formalization}

Let $x\in X$ ({\em e.g.}, an image) and $y \in Y = \{y_1, \dots, y_K\}$ ({\em e.g.}, a label), identified with an integer in $\{1, \dots, K\}$. Let $p$ denote an unknown distribution from which a dataset is drawn ${\cal D} = \{(x_i, y_i) \sim p(x,y)\}_{i=1}^N$. Let $p^{\textnormal{old}}$ be the pseudo-distribution associated with the old ({\em reference}) model, and $p^{\textnormal{new}}$ the same for the new one.\footnote{In addition to having different architectures, the models can be trained with different optimization schemes, minimizing different loss functions, on different datasets. For simplicity, below we consider the case of models with same or different architectures, trained on the same dataset ${\cal D}$ using the same loss function and optimization scheme.
} These are parametric functions that, ideally, approximate the true posterior which is the optimal (Bayesian) discriminant, in the sense of minimizing the expected probability of error. More specifically, when evaluated on the sample $(x_i, y_i)$, $p^{\textnormal{new}}(\cdot | \cdot)$ takes the form
\begin{equation}
    p^{\textnormal{new}}(y = y_i| x = x_i) = \frac{\exp(\langle \vec{y}_i, \phi_w^{\textnormal{new}}(x_i) \rangle)}{\sum_{j=1}^K \exp(\langle \vec{y}_j, \phi_w^{\textnormal{new}}(x_i) \rangle)}
\label{eq:softmax}\end{equation}
where $\vec{y}_i \in \mathbb{R}_{\ge 0}^K$ is the ``one-hot'' (indicator) vector corresponding to the categorical variable $y_i \in \{1, \dots, K\}$. $p^{\textnormal{old}}$ has a similar form but with a different {\em architecture} $\phi$ that has known parameters.  We refer to $\phi_w$ as the {\em model} or {\em discriminant}. 
The final class prediction of the model, $ \phi_w $, is denoted by $ \hat{y} $, where, $ \hat{y} \parens{x_i} = \argmax_{y} p \parens{ y | x_i } $.

\subsection{Negative Flips}
\label{sec:neg_flips}

In Fig.~\ref{fig:teaser} we illustrate the different errors occuring after a model upgrade, comparing the predictions $ \hat{y}^{\textnormal{old}} $ and $ \hat{y}^{\textnormal{new}} $. The consistent areas are where  $ \phi_w^{\textnormal{old}} $ and $ \phi_w^{\textnormal{new}} $ are both either correct ($\hat{y}^{\textnormal{old}}(x_i) = \hat{y}^{\textnormal{new}}(x_i) = y_i$) or incorrect ($\hat{y}^{\textnormal{old}}(x_i) \neq y_i, \hat{y}^{\textnormal{new}}(x_i) \neq y_i$). Positive flips occur when samples were incorrectly predicted by $ \phi_w^{\textnormal{old}} $, but correctly predicted by $ \phi_w^{\textnormal{new}} $. The most relevant to this paper are  \textbf{negative flips}, where samples flipped from  correct $ \phi_w^{\textnormal{old}} $ to incorrect predictions  $ \phi_w^{\textnormal{new}} $. The {\bf negative flip rate (NFR)} measures the fraction of samples that are negatives
\begin{equation}
    \textnormal{NFR} = \dfrac{1}{N} \sum_{i = 1}^{N} \mathbf{1}(\hat{y}_{i}^{\text{new}} \neq y_i,  \hat{y}_{i}^{\text{old}} = y_i)
\end{equation}
where, $ \mathbf{1}(\cdot)$ is the indicator function. 

\subsection{Persistence of Negative Flips}
\label{sec:existence_regression}

Deep neural network (DNN) classifiers are usually trained by minimizing the empirical cross-entropy (CE) loss:
\begin{equation}
\min_w \mathcal{L}_{\textnormal{CE}} (\phi, w) = \min_w \frac{1}{N} \sum_{i=1}^{N}- \log p_{w} \parens{ y_i | x_i } 
\label{eq:standard_loss_fn}
\end{equation}
This objective is minimized by Stochastic Gradient Descent (SGD)~\cite{kiefer1952sgd} annealed to convergence to or near a local minimum.\footnote{In reality, the landscape of DNNs trained with CE does not comprise isolated (Morse) critical points, but rather ``wide valleys'' \cite{choromanska2015loss,chaudhari2019entropy}, and convergence is only forced by annealing the learning rate, without which the solution travels on limit cycles.
} The final discriminant is determined by (a) the DNN architecture  $ \phi $, (b) the training methodology, including optimization scheme and associated  hyper-parameters such as learning rate and momentum, ($ \eta, \mu $), (c) the dataset on which the model is trained $ \cal D $ and (d) the initialization $ w^{(0)} $. 
Unless explicitly enforced, a new model is typically not ``close'' to an old one. Even if based on the same architecture and trained on the same dataset, different runs on SGD can converge to distant points in the loss landscape \cite{chaudhari2018stochastic}. What is similar along limit cycles of solutions is the average error rate in the training set, which is minimal; what is different is the samples on which errors occur. These are negative flips, measured empirically in Fig. \ref{fig:acc_diff_vs_nfr}. There, we measure the difference of error rates and NFR between pairs of models trained on the ILSVRC12~\cite{ILSVRC15} dataset\footnote{All models in this experiment are publicly available in the PyTorch~\cite{paszke2019pytorch} online model zoo}.
We observe that, from earlier architecture design such as AlexNet~\cite{krizhevsky2017imagenet}, to recent ones such as DenseNet~\cite{huang2017densely}, although the overall error rates have dropped, the NFR remain non-negligible. In many cases, the NFR is of the same order of magnitude as the accuracy gain from the model update. This stubborn phenomenon  calls for a dedicated solution that does not rely on reducing the error rate to zero in order to have regression-free model updates.

\begin{table*}[ht]
    \centering
    \begin{tabular}{|c||c|c|c|}
        \hline
        Approach & $ \mathcal{F}$ & $\mathcal{D}$ & $q$ \\
        \hline 
        Naive  Basline & $ \mathbf{1}(\hat y^{\textnormal{old}}(x_i) = y_i) $ & Cross Entropy & $\vec{y}_i$ \\
        \hline
        Focal Distillation - KL (FD-KL) & $ \alpha + \beta \cdot \mathbf{1}(\hat{y}^{\textnormal{old}}(x_i) = y_i) $ & $\tau$-scaled KL-Divergence& $ p^{\textnormal{old}}(y|x_i)$ \\
        \hline
        Focal Distillation - Logit Matching (FD-LM) & $ \alpha + \beta \cdot \mathbf{1}(\hat{y}^{\textnormal{old}}(x_i) = y_i) $ & $\ell_2$ distance & $ \phi_{w}^{\textnormal{old}}(x_i) $ \\
        \hline
    \end{tabular}
    \caption{Different approaches for targeted positive congruent training and the corresponding choices of $\mathcal{F}$, $\mathcal{D}$, and $q$ in the PC loss functions.  }
    \label{tab:pc_loss_design}
\end{table*}

\section{Positive Congruent Training}\label{sec:pct}

In most applications, the reference model is inherited and  cannot be changed directly, so PC training is limited to the new model. However, in some cases we may get to design the reference model,  aiming to make it easily updated with PCT.  For the first  ``backward'' case,  we first present a simple approach to reducing NFR and achieve  PCT. The idea is to simultaneously maximize the  ``both correct'' region in Fig.~\ref{fig:teaser}, which measures \emph{positive congruence}, along with the overall accuracy of the new model on the training set.

To achieve PC Training, we use the following objective function
\begin{equation}
    \min_w \mathcal{L}_{\textnormal{CE}} \parens{\phi^{\textnormal{new}}, w} + \lambda \mathcal{L}_{\textnormal{PC}} (\phi^{\textnormal{new}}, w;  \phi^{\textnormal{old}} )
\label{eq:pc-training}\end{equation}
where $\mathcal{L}_{\textnormal{CE}}$ is defined in (\ref{eq:standard_loss_fn}) and  $\mathcal{L}_{\textnormal{PC}} (\phi^{\textnormal{new}}, \phi^{\textnormal{old}}, w)$ is  the \emph{positive congruence (PC) loss} with multiplier $\lambda$. We propose a generic form of the PC loss function as
\begin{equation}
\lambda \mathcal{L}_{\textnormal{PC}} (\phi^{\textnormal{new}}, w;  \phi^{\textnormal{old}} ) = \mathcal{F}(x_i)\mathcal{D}(\phi_{w}^{\textnormal{new}}(x_i), q(x_i)),
\label{eq:pc_loss}\end{equation}
where $\mathcal{F}$ is a filter function $\mathcal{F} \in f:X \rightarrow \mathbb{R}_{\ge 0} $ that applies a weight for each training sample based on the model outputs and $\mathcal{D}$ is a distance function that measures the difference of the new model's outputs to a certain target vector $q(x_i)$ conditioned on $x_i$.  
The PC loss serves to bias training towards maximal positive congruence. Favoring iso-error-rate solutions \cite{chaudhari2018stochastic} with  maximal positive congruence can lead to lower NFR relative to the reference model.    
In Table~\ref{tab:pc_loss_design} we illustrate the different choices of $\mathcal{F}$ and $\mathcal{D}$. We introduce different PC losses in the the next two sections.

\subsection{Naive Baseline}
\label{sec:baselines}
We first consider a simple PC loss that measures CE on the samples where the old model made correct predictions, denoted as
\begin{equation}
    \mathcal{L}_{\textnormal{PC}}^{\textnormal{naive}} = \frac{1}{N} \sum_{i=1}^N - \mathbf{1}(\hat y^{\textnormal{old}}(x_i) = y_i) \log p_w^{\textnormal{new}}(y_i | x_i).
    \label{eq:naive}
\end{equation}
This is just a re-weighting of samples that the old models classified correctly by a factor $1 + \lambda$. The resulting model is labeled ``Naive'' in the experiments in Sect. \ref{sec:experiments}. 
This method does not help reduce the NFR: It is hard to find a suitable hyperparameter $\lambda$ that gives sufficient weights to positive samples without inducing rote memorization. We now explore a more direct way to pass information from the old model to the new one.

\subsection{Focal Distillation}
\label{sec:distillation}

Knowledge distillation~\cite{hinton2015distilling} aims at biasing the new model to be ``close'' to the old one during training. In this sense it has the potential to reduce the NFR, but also to reduce positive flips, which are instead beneficial. Focal Distillation (FD) mitigates this risk by using the loss
\begin{eqnarray}
    \mathcal{L}_{\textnormal{PC}}^{\textnormal{Focal}} = -\sum_{i = 1}^{N} [\alpha + \beta \cdot \mathbf{1}(\hat{y}^{\textnormal{old}}(x_i) = y_i)] \mathcal{D}(\phi_{w}^{\textnormal{new}}, \phi_{w}^{\textnormal{old}}),
\label{eq:distll}\end{eqnarray}
which penalizes a distance $\mathcal{D}$  between the output of the two models,  weighted by a filtering function $\mathcal{F}=\alpha + \beta \cdot \mathbf{1}(\hat{y}^{\textnormal{old}}(x_i) = y_i)$, to the target  $q = \phi_{w}^{\textnormal{old}}(x_i)$. The filter function applies a basic weight $\alpha$ for all samples in the training set and an additional weight to the samples correctly predicted by the old model. 
This biases the new model towards the old one for positive samples in the old model, thus imposing a cost on NFR as well as the overall error rate. 
When $\alpha=1$ and $\beta=0$, focal distillation reduces to ordinary distillation~\cite{hinton2015distilling}. 
When $\alpha = 0$ and $\beta>0$, we are only applying the distillation objective to the training samples predicted correctly by the old model. We assess these choices empirically in Sec.~\ref{sec:simple_case}.

A possible choice of $\mathcal{D}$ is the temperature scaled KL-divergence of~\cite{hinton2015distilling}:
\begin{equation}
    \mathcal{D}^{\mathrm{KL}}(\phi_{w}^{\textnormal{new}}, \phi_{w}^{\textnormal{old}}) = \mathrm{KL}\brackets{\sigma(\frac{\phi_{w}^{\textnormal{new}}(x_i)}{\tau}), \sigma(\frac{\phi_{w}^{\textnormal{old}}(x_i)}{\tau})}.
\label{eq:KD}\end{equation}
Here, $\sigma$ is the ``Softmax'' function as shown in Eq.~(\ref{eq:softmax}) and $\tau$ is the temperature scaling factor. 
In Sec.~\ref{sec:distillation}, we found that $\tau$ has to be set to a large number (\emph{e.g.} $\tau=100$) for this PC loss to reduce NFR.
In this case, the above approaches the distance among  ``logits''~\cite{hinton2015distilling,Bucila2006ModelC}
\begin{equation}
    \mathcal{D}^{\mathrm{LM}}(\phi_{w}^{\textnormal{new}}, \phi_{w}^{\textnormal{old}}) = \frac{1}{2} \|\phi_{w}^{\textnormal{new}}(x_i) - \phi_{w}^{\textnormal{old}}(x_i)\|_2^2.
\label{eq:logit_matching}\end{equation}
In Sect. \ref{sec:experiments}  we compare various settings of FD with the naive baseline and observe that training with the FD can indeed reduce the NFR at the cost of a slight increase n average accuracy for the new model.

\subsection{PC Training by Ensembles}
\label{sec:ensembles}

The fact that there are iso-error sets in the loss landscape of DNN  (loci in weight space that correspond to models with equal error rate${}^3$) \cite{chaudhari2018stochastic} can be used to our advantage to reduce NFR as outlined in the introduction. In some cases, one may be able to choose both the new {\em and the old models} in a manner that reduces the NFR. 
Geometrically, one can think of each model as representing an iso-error equivalence class: A sphere in the space of models, centered at zero error, where all models achieve the same error rate but differ by which samples they mistake. Since cordal averaging reduces the distance to the origin \cite{watson1983statistics} and therefore the average distance to other models (for instance, future PC trained ones), we hypothesize that replacing each model with an ensemble trained on the same data might reduce the NFR.  

In particular, we assume that $ \phi^{\textnormal{old}} $ and $ \phi^{\textnormal{new}} $ are trained independently, each from a collection of models $ \{ \phi^{\textnormal{old}}_j \}_{j = 1}^N $ and $ \{ \phi^{\textnormal{new}}_j \}_{j = 1}^N $. We combine the results from all the models by averaging their discriminants
\begin{equation}
    p^{\textnormal{ensemble}} \parens{ y |x_i } = \sigma\brackets{\frac{1}{L} \sum_{j = 1}^L \phi_{w_j}^{\textnormal{old}}\parens{y|x_i}}.
    \label{eq:ensemble}
\end{equation}
where $L$ is the number of models in the ensemble. Similarly, we can define $ \phi^{\textnormal{new}} $. Note that models in each ensemble are trained independently with different initialization.

We test our hypothesis empirically in Sect. \ref{sec:experiments} where we find that, indeed, the NFR between the two ensembles is lower than that between any two individual models. 

While this may not seem surprising at first, since ensembles reduce error rate,
we note that, as anticipated in the introduction, NFR can be reduced independently of ER. In fact, empirically we observe that {\em the NFR decreases more rapidly than the average error} as the size of the ensemble grows (Fig.~\ref{fig:ensemble_nfr_change_arch}), indicating that ensembling improves PC training beyond simply lowering the average error rate. Note the this phenomenon holds for any pair of ensembles with either the same or different architectures. This makes it suitable as a paragon to explore the upper limit of PC training. 
While test-time ensembles are not practical at scale, this observation points to promising areas of future investigation by collapsing ensembles.

\section{Experiments}
\label{sec:experiments}

We test the methods presented on image classification tasks using ImageNet~\cite{deng2009imagenet} and iNaturalist~\cite{van2018inaturalist}. 
We start with the simplest case: same architecture and training data, different training runs. We compare baseline methods for PCT and emsembles. We then extend the approaches to encompass  1) architecture changes; 2) changing number of training samples per class; 3) increase in the number of classes. Finally, we present empirical evidence on the source of negative flips and how  PC training affects them. \\

\noindent{\bf Relative NFR.} Since overall error rate (ER) is an upper bound to the NFR, comparison is challenging across datasets with different ERs. For this reason, we introduce the relative NFR: \begin{equation}
    \textnormal{NFR}_{\textnormal{rel}} = \frac{\textnormal{NFR}}{(1 - \textnormal{ER}_{\textnormal{old}}) * \textnormal{ER}_{\textnormal{new}}},
\label{eq:rel_nfr}\end{equation}
where $\textnormal{ER}_{\textnormal{new}}$ and $\textnormal{ER}_{\textnormal{old}}$ denote the error rate of $\phi_{\textnormal{new}}$ and $\phi_{\textnormal{old}}$.
The denominator is the expected error rate on the subset of samples predicted correctly by the old model. This is a naive estimation of NFR if the two models are independent of each other. The relative NFR is a measure of reduction in regression from a PCT method which factors out overall model accuracy.

\subsection{Implementation Details}

Unless otherwise noted, we minimize the CE loss~(\ref{eq:standard_loss_fn}) with batch-size  $2048$ for $90$ epochs on each dataset. The learning rate starts at $0.1$ and decreases by $1/10$ every $30$ epochs. We use $\lambda=1$ for Eq.~\ref{eq:pc-training}.  Focal distillation and ensembles  are implemented with  PyTorch~\cite{paszke2019pytorch}. 
Models in Fig.~\ref{fig:acc_diff_vs_nfr} are from the PyTorch model zoo.

\begin{table}[t]
\small
\centering
\begin{subtable}[h]{\linewidth}
\setlength{\tabcolsep}{1.5mm}{
\begin{tabular}{l||c|c|c|c|c}
\hline
\multirow{2}{*}{PCT Approach}        & \multicolumn{2}{c|}{Error Rate~(\%)} & \multirow{1}{*}{NFR} & \multirow{1}{*}{Rel. NFR} & \multirow{2}{*}{\#Params} \\
\cline{2-3}
& $\phi^{\textnormal{old}}$ & $\phi^{\textnormal{new}}$ &(\%)&(\%) \\
\hline 
No Treatment            &30.24&30.29&6.44&30.48&12M  \\ \hline
Naive                   &30.24&29.34&5.72&27.95&12M  \\ \hline
BCT~\cite{bct}                    &30.24&29.66&6.39&30.88&12M \\ \hline
FD-KL                   &30.24&30.63&2.50&11.70&12M  \\ \hline
FD-LM                   &30.24&30.47&2.35&11.06&12M  \\ \hline\hline
Ensemble                &26.07&25.98&1.70&8.85&187M \\ \hline
\end{tabular}
}
\caption{ILSVRC12}\label{tab:sample_exp_ilsvrc12}
\end{subtable}
\hfill
\begin{subtable}[h]{\linewidth}
\setlength{\tabcolsep}{1.5mm}{
\begin{tabular}{l||c|c|c|c|c}
\hline
\multirow{2}{*}{PCT Approach}        & \multicolumn{2}{c|}{Error Rate~(\%)} & \multirow{1}{*}{NFR} & \multirow{1}{*}{Rel. NFR} & \multirow{2}{*}{\#Params} \\
\cline{2-3}
& $\phi^{\textnormal{old}}$ & $\phi^{\textnormal{new}}$ &(\%)&(\%) \\
\hline 
No Treatment    & 40.69&	41.06&	7.77&	31.91&	14M \\ \hline
Naive           & 40.69&	45.47&	11.07&	41.05&	14M  \\ \hline
FD-KL           & 40.69&	41.81&	2.83&	11.41&	14M  \\ \hline
FD-LM           & 40.69&	41.78&	2.71&	10.94&	14M  \\ \hline \hline
Ensemble     & 35.68&	35.42&	2.01&	8.82&	221M  \\ \hline
\end{tabular}}
\caption{iNaturalist}\label{tab:sample_exp_inaturalist}
\end{subtable}
\caption{Experiments on training the same model architecture multiple times on three datasets: (a) ILSVRC12~\cite{ILSVRC15}, (b) iNaturalist~\cite{van2018inaturalist}. Applying PC training in the case can significantly reduce the NFR without hurting the new model's accuracy. Among them, the focal distillation works the best for the backward compatibility setting. ``Ensemble'' refers to PC training with ensembles in the forward setting. ``Rel. NFR'' refers to the relative NFR value shown in~Eq.\ref{eq:rel_nfr}. ``\#Params'' refers to the number of parameters of the new model.  }\label{tab:same_exp}
\end{table}

\subsection{Positive Congruent Training}\label{sec:simple_case}
We start with the same architecture~\cite{he2016resnet} trained twice on the same dataset, either ILSVRC12~\cite{ILSVRC15} or iNaturalist~\cite{van2018inaturalist},  and report results on their official validation sets in Table~\ref{tab:same_exp}\footnote{We follow the common practice to use the official validation sets on these datasets as test set and a small portion training set for validation.}.

\noindent{\bf Variants of PCT} include
the {\bf naive baseline} in Eq.~\ref{eq:naive}, which unfortunately does not reduce the NFR markedly. Yet it is worth noting that, due to the fact that models are not independent, being trained on the same data, their relative NFR are always less than $100\%$. We additionally evaluate the recently proposed BCT method~\cite{bct} for aligning representation between models. It does not reduce NFR.

{\bf Focal distillation} in Eq.~\ref{eq:distll} (FD) has two variants, either using the KL-divergence after soft-max (\emph{FD-KL}) or matching logits before soft-max (\emph{FD-LM}).  In FD,  there are two parameters, $\alpha$ and $\beta$, controlling the filter function $\mathcal{F}$. We set $\alpha=1$ and $\beta=5$ in this set of experiments. The results on ILSVRC12 are summarized in Table~\ref{tab:sample_exp_ilsvrc12}. We see that training with both variants of focal distillation, we can reduce the NFR from $6\%$ to around $2.5\%$, a $60\%$ relative reduction. We also note that the reduction of NFR comes as the cost of a slightly increased error rate of the new model.

Finally we evaluate the {\bf ensemble} approach. We use two independently trained ensembles each composed of $16$ ResNet-18~\cite{he2016resnet} for $\phi^{\textnormal{old}}$ and $\phi^{\textnormal{old}}$. We see ensembles achieve the lowest absolute NFR, as low as $1.7\%$. Even when discounting the accuracy gain of ensembles, we also observe better relative NFR than other approaches in the backward setting. The price to pay is computational cost, a multiplier depending on the number of models in the ensemble.

\noindent{\bf Training from scratch vs fine-tuning.}
In practical scenarios, we often fine-tune from an existing model pretrained on a large-scale dataset. To compare the impact of this, on the ImageNet dataset we trained all models from scratch but on the iNaturalist dataset~\cite{van2018inaturalist} we fine-tune from model weights pretrained on ImageNet. As shown in Table~\ref{tab:sample_exp_inaturalist}, we observe starting from a common pretrained model does not guarantee regression-free. We observe a $7\%$ NFR without treatment in finetuning on iNaturalist. But with PC training, we can reduce the NFR to $2.01\%$

\noindent{\bf Distances in focal distillation}
In focal distillation we presented two different choices of distance. In~Table~\ref{tab:same_exp}, we observe that the logit matching distance function (FD-LM) leads to marginally better NFR reduction. It is also worth to note that the KL-divergence only effectively reduces NFR when the temperature scaling parameter $\tau$ becomes  large, say $100$,  at which point it resembles logit matching~\cite{hinton2015distilling}.

\begin{table}[t]
\small
\centering
\begin{tabular}{c|c|c|c|c}
\hline
$\alpha$ & $\beta$ & $\phi^{\textnormal{new}}$ Error Rate ($\%$) & NFR $(\%)$ & Rel. NFR$(\%)$ \\ \hline
0        & 0       & 30.29                                       & 6.44       & 30.48          \\ \hline
0        & 1       & 31.52                                       & 5.25       & 23.88          \\ \hline
1        & 0       & 31.12                                       & 3.90       & 17.96          \\ \hline
1        & 1       & 30.59                                       & 2.75       & 12.89          \\ \hline
1        & 2       & 30.79                                       & 2.73       & 12.71          \\ \hline
1        & 5       & 30.47                                       & 2.35       & 11.06          \\ \hline
1        & 10      & 30.44                                       & 2.39       & 11.26          \\ \hline
1        & 20      & 33.55                                       & 6.94       & 29.65          \\ \hline
\end{tabular}
\caption{Effect of different $\alpha$ and $\beta$ values in focal distillation. We train ResNet-18 for both $\phi^{\textnormal{old}}$ and $\phi^{\textnormal{new}}$ on ILSVRC12 dataset~\cite{ILSVRC15}. We use the logit matching distance function for focal distillation because it has better effect in reducing NFR. }\label{tab:focal_distill}
\end{table}

\noindent{\bf Weights and focus in focal distillation}
In Focal Distillation, we use a default focal weight  $\beta=5$ and a base weight $\alpha=1$. We now experiment changing the focal weight and base weight and report results in Table~\ref{tab:focal_distill}. 
First note two special cases: 1) $\alpha=0$ and $\beta=1$, when we only apply distillation to samples classified correctly by the old model, which is ineffective: The new model must still learn the old model's behavior from all samples in order to be able to overcome negative flips. 2) $\alpha=1$ and $\beta=0$, corresponding to ordinary  distillation, which can help reduce NFR only to a limited extent. Focal Distillation is most effective when $\alpha=1$ and $\beta$ is between $5$ to $10$, reducing NFR to $2.35\%$.

\begin{figure}[t]
    \centering
    \begin{subfigure}[b]{0.9\linewidth}
         \centering
         \includegraphics[width=1.1\linewidth]{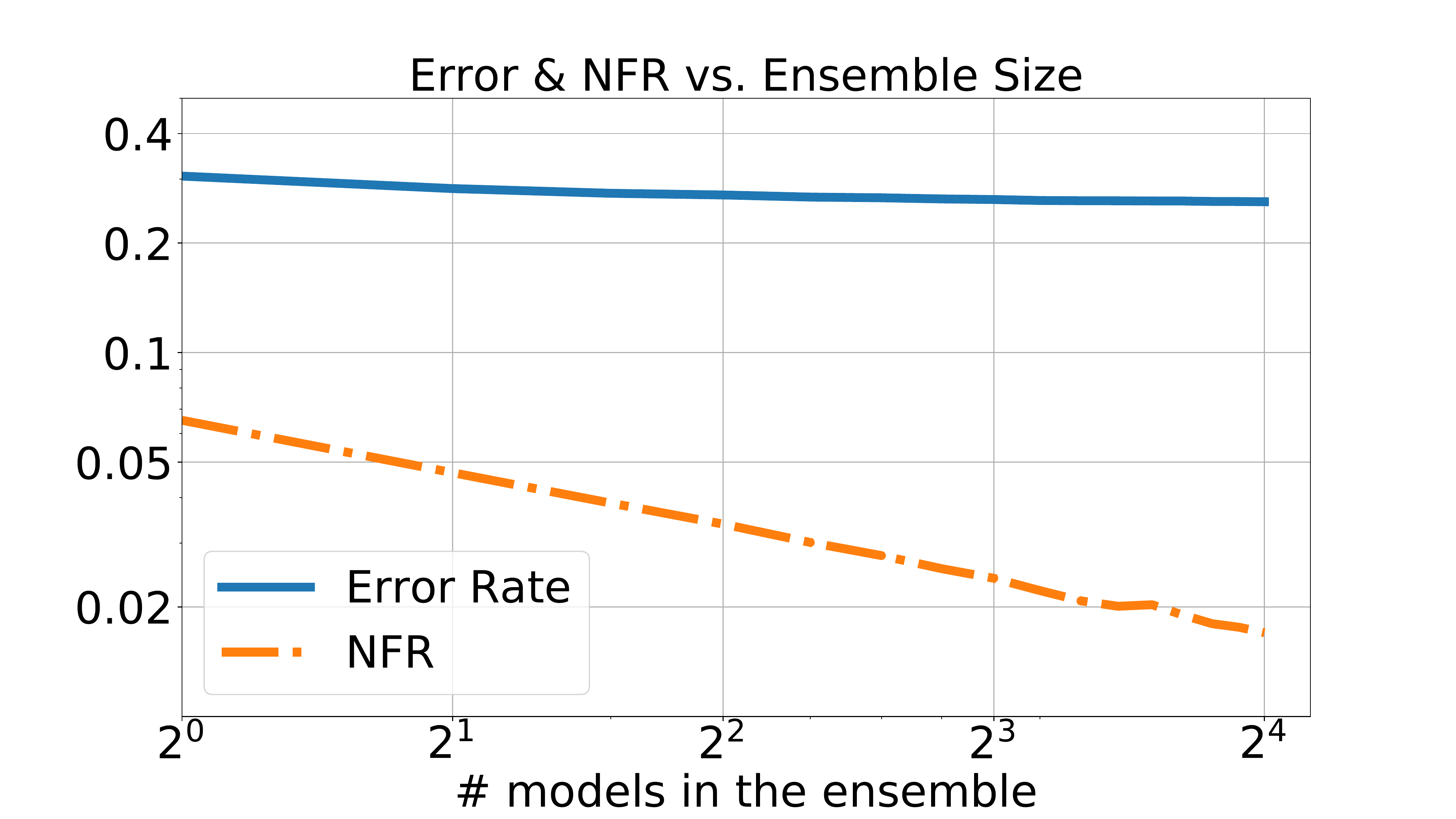}
         \caption{ResNet-18 $\rightarrow$ ResNet-18}
         \label{fig:ensemble_nfr_change_arch:a}
     \end{subfigure}
     \hfill
     \begin{subfigure}[b]{0.9\linewidth}
         \centering
         \includegraphics[width=1.1\linewidth]{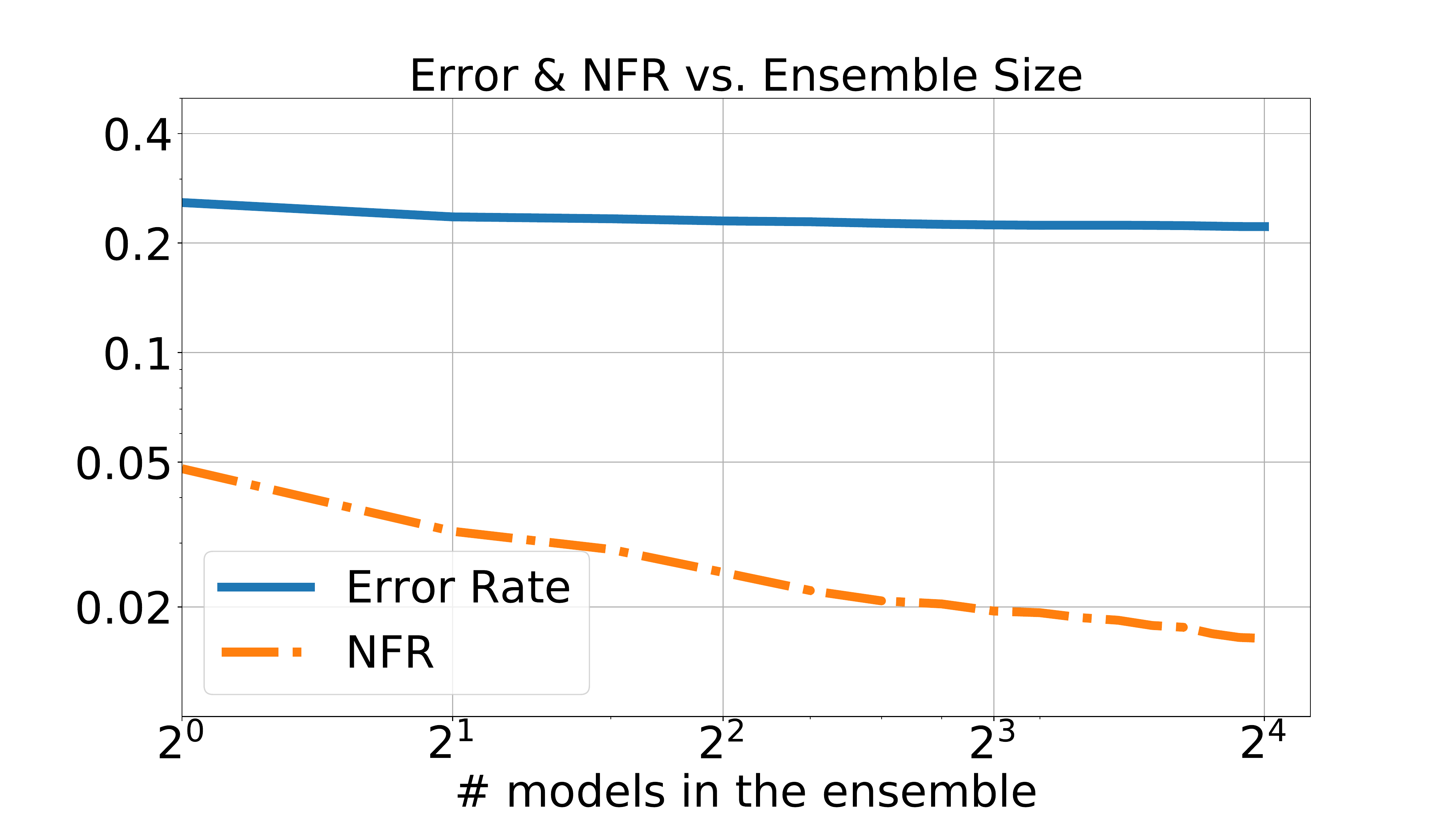}
         \caption{ResNet-18 $\rightarrow$ ResNet-50}
         \label{fig:ensemble_nfr_change_arch:b}
     \end{subfigure}
    \caption{The effect of ensemble size (number of individual models) on error rates and the NFRs when both the old and new models are both independently trained ensembles. Results are reported on ILSVRC12. (a) shows the results when the old model and new one are both composed of ResNet-18~\cite{he2016resnet} models. (b) shows the case when the old model is an ensemble of ResNet-18~\cite{he2016resnet} models and the new one is formed by ResNet-50~\cite{he2016resnet} models. In both case, the error rates plateau quickly but the NFRs keep decreasing as the ensemble size increases.  }
    \label{fig:ensemble_nfr_change_arch}
\end{figure}

\noindent{\bf Impact of ensemble size}
In ensemble experiments on ILSVRC12, we used a default ensemble size of $16$. In Figure~\ref{fig:ensemble_nfr_change_arch} we explore the behavior of NFR as the number of  components in an ensemble changes. Although the accuracy plateaus with the growing ensemble size, the NFR keeps decreasing in a logarithmic scale. This observation is also confirmed in the case of ensembles with different architectures. Although having a large ensemble is impractical for real-world applications, these results corroborate the geometric picture sketched in Sect.~\ref{sec:ensembles} and suggest that it may  be possible to achieve zero NFR in a model update without having to drive the ER to zero, which may be infeasible, as suggested in the Introduction.

\subsection{Other Changes in Model Updates}
Having studied the simplest case of PC training, we now move to more practical cases of image classification model updates. We enumerate and study the following types of changes: 1) model architecture; 2) number of training samples per class; 3) number of classes.

\noindent{\bf Changing model architectures}
to a larger model, with either a similar architecture (Resnet-18 to Resnet-50) or a dissimilar one (ResNet18 to DenseNet-161) , yields results summarized in Table~\ref{tab:arch_change}. A larger model is expected to have a lower overall error rate. However, without any treatment, it still suffers from a significant number of negative flips. 
We then apply the focal distillation approach in this case, we see a redution in NFR, although the new model incurs a slight increase in error rate.
We also observe that ensembles reduce NFR, this time without increasing the new model's error rate.
This is noteworthy because the two ensembles in this case do not share any information other than the dataset on which they are trained.
In Fig.~\ref{fig:ensemble_nfr_change_arch} we also visualize the trend of NFR vs. the ensemble size. It suggests that, even in the case of architecture changes, it may be possible to achieve zero NFR.

\noindent{\bf Increasing  training samples}
per visual category is shown in Table~\ref{tab:data_growth} for  the ILSVRC12 dataset, with the old model trained on all 1000 classes using $50\%$ of the samples per category, and the new model trained on $100\%$ set. We observe that all new models decrease the ER, but PC trained ones achieve better reduction of NFR.

\noindent{\bf Increasing the number of classes} is tested on ILSVRC12 with the old model trained on a random subset of $500$ classes and the new model on the entire ILSVRC12 training set. We use ResNet-50  for this study. The evaluation is done on the validation subset which comprises all samples of the original 500 classes on which the older model is trained on. The results are shown in Table~\ref{tab:class_growth}.
Since the number of categories increase for the newer model, so does the total error on the validation set. Yet, PC training is able to reduce NFR.

\begin{table}[t]
\small
\centering
\begin{subtable}[h]{\linewidth}
\setlength{\tabcolsep}{1.5mm}{
\begin{tabular}{l||c|c|c|c|c}
\hline
\multirow{2}{*}{Approach}        & \multicolumn{2}{c|}{Error Rate~(\%)} & \multirow{1}{*}{NFR} & \multirow{1}{*}{Rel. NFR} & \multirow{2}{*}{\#params} \\
\cline{2-3}
& $\phi^{\textnormal{old}}$ & $\phi^{\textnormal{new}}$ &(\%)&(\%) \\
\hline 
No Treatment            &30.24&25.85&4.89&27.12&25M    \\ \hline
Naive      &30.24&24.41&3.78&22.20&25M  \\ \hline
FD-KD                 &30.24&26.32&2.90&15.79&25M  \\ \hline
FD-LM                   &30.24&26.53&2.92&15.78&25M  \\ \hline\hline
Ensemble           &26.07&22.20&1.64&9.99&409M \\ \hline
\end{tabular}}
\caption{ResNet-18 $\rightarrow$ ResNet-50}
 \end{subtable}\hfill
\begin{subtable}[h]{\linewidth}
 \setlength{\tabcolsep}{1.5mm}{
 \begin{tabular}{l||c|c|c|c|c}
 \hline
 \multirow{2}{*}{Approach}        & \multicolumn{2}{c|}{Error Rate~(\%)} & \multirow{1}{*}{NFR} & \multirow{1}{*}{Rel. NFR} & \multirow{2}{*}{\#params} \\
 \cline{2-3}
 & $\phi^{\textnormal{old}}$ & $\phi^{\textnormal{new}}$ &(\%)&(\%) \\
 \hline 
 No Treatment            &30.24&22.86&4.03&25.24&25M    \\ \hline
 Naive                   &30.24&21.60&3.28&21.76&25M  \\ \hline
 FD-KD                   &30.24&23.52&2.50&15.24&25M  \\ \hline
 FD-LM                   &30.24&23.83&2.56&15.40&25M  \\ \hline
 % \hline
 % Ensemble (4)                &28.02&22.74&2.41&14.72&100M \\ \hline
 \end{tabular}}
 \caption{ResNet-18 $\rightarrow$ DenseNet161}
 \end{subtable}
\caption{Experiments of PC training methods in changes of model architectures on ILSVRC12~\cite{ILSVRC15}. Here the old model architecture is ResNet-18~\cite{he2016resnet}. We experiment with the new models with both Resnet-50~\cite{he2016resnet} and Denset161~\cite{huang2017densely} architectures. Results suggest that PC training method are effective in reducing regression in the face of model architecture changes. }\label{tab:arch_change}
\end{table}

\begin{table}[t]
\small
\centering
\begin{subtable}[h]{\linewidth}
\setlength{\tabcolsep}{1.5mm}{
\begin{tabular}{l||c|c|c|c|c}
\hline
\multirow{2}{*}{Approach}        & \multicolumn{2}{c|}{Error Rate~(\%)} & \multirow{1}{*}{NFR} & \multirow{1}{*}{Rel. NFR} & \multirow{2}{*}{\#params} \\
\cline{2-3}
& $\phi^{\textnormal{old}}$ & $\phi^{\textnormal{new}}$ &(\%)&(\%) \\
\hline 
No Treatment            &    28.58   &    23.65     &  3.98   &  19.47  & 25M     \\ \hline
Naive                   &    28.45   &    24.46     &  3.27   &  18.68  & 25M     \\ \hline
FD-KD                   &    28.45   &    25.20     &  2.89   &  15.84  & 25M     \\ \hline
FD-LM                   &    28.45   &    24.77     &  2.85   &  16.09 & 25M     \\ \hline\hline
Ensemble                &    26.39   &    22.09     &  2.75   &  16.88  & 100M    \\ \hline
\end{tabular}}
\caption{Increase in \# samples}\label{tab:data_growth}
\end{subtable}\hfill
\begin{subtable}[h]{\linewidth}
\setlength{\tabcolsep}{1.5mm}{
\begin{tabular}{l||c|c|c|c|c}
\hline
\multirow{2}{*}{Approach}        & \multicolumn{2}{c|}{Error Rate~(\%)} & \multirow{1}{*}{NFR} & \multirow{1}{*}{Rel. NFR} & \multirow{2}{*}{\#params} \\
\cline{2-3}
& $\phi^{\textnormal{old}}$ & $\phi^{\textnormal{new}}$ &(\%)&(\%) \\
\hline 
No Treatment         &    19.30   &    23.65     &  8.05   &  41.90  & 25M     \\ \hline
Naive                &    19.28   &    23.74     &  7.70   &  40.20  & 25M     \\ \hline
FD-KD                &    19.28   &    24.14     &  7.07   &  36.29  & 25M     \\ \hline
FD-LM                &    19.28   &    25.11     &  7.37   &  36.35  & 25M     \\ \hline\hline
Ensemble             &    17.53   &    21.98     &  6.72   &  37.06  & 100M    \\ \hline
\end{tabular}}
\caption{Increase in \# classes}\label{tab:class_growth}
\end{subtable}
\caption{Experiments of training data change on ILSVRC12~\cite{ILSVRC15} dataset. We use Renset-18 architecture for both $\phi^{\textnormal{old}}$ and $\phi^{\textnormal{old}}$. }\label{tab:data_change}
\end{table}

\begin{figure}[th]
    \centering
    \begin{subfigure}[b]{.88\linewidth}
         \centering
         \includegraphics[width=1.07\linewidth]{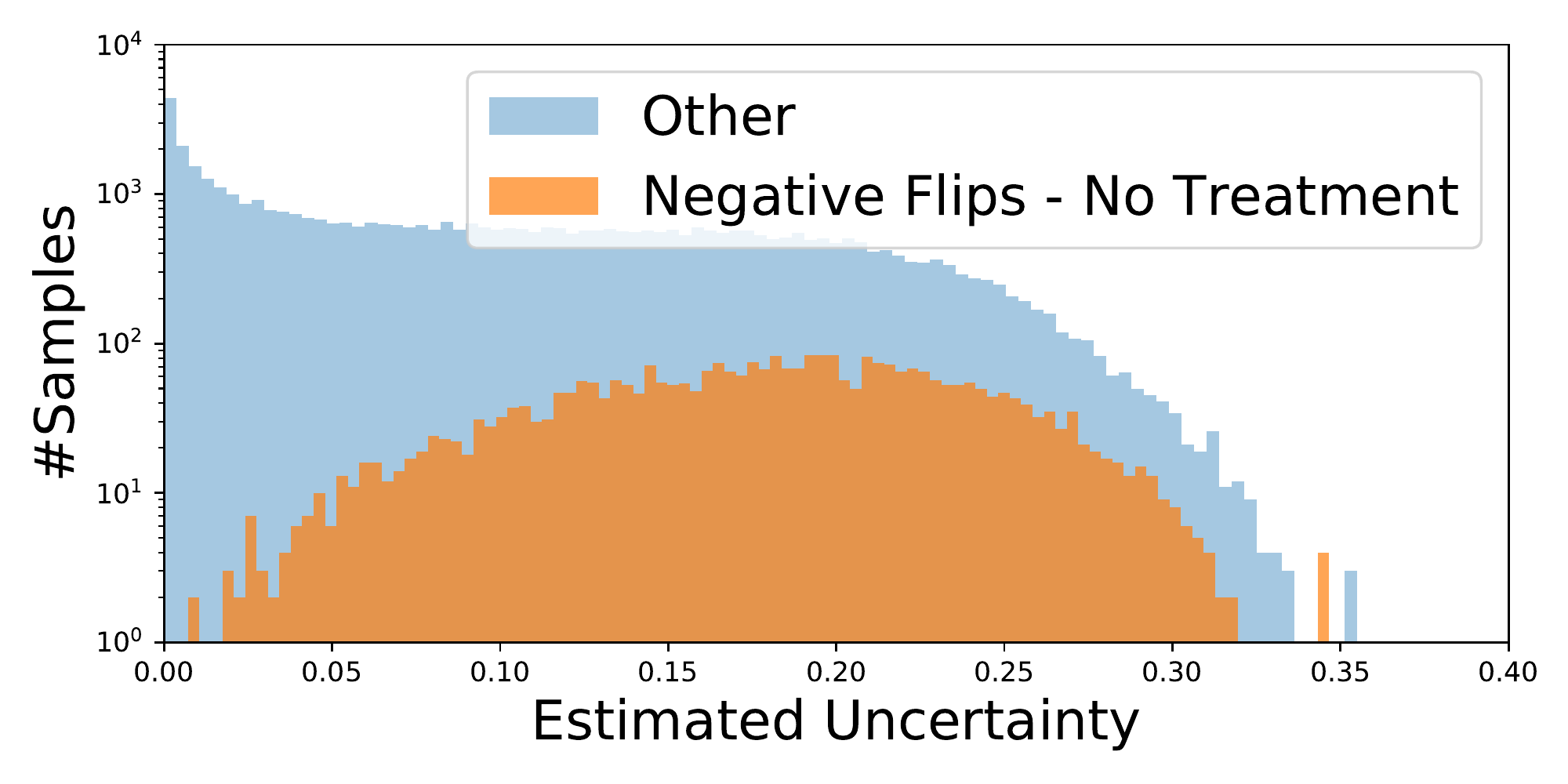}
     \end{subfigure}
     \hfill
     \begin{subfigure}[b]{.88\linewidth}
         \centering
         \includegraphics[width=1.07\linewidth]{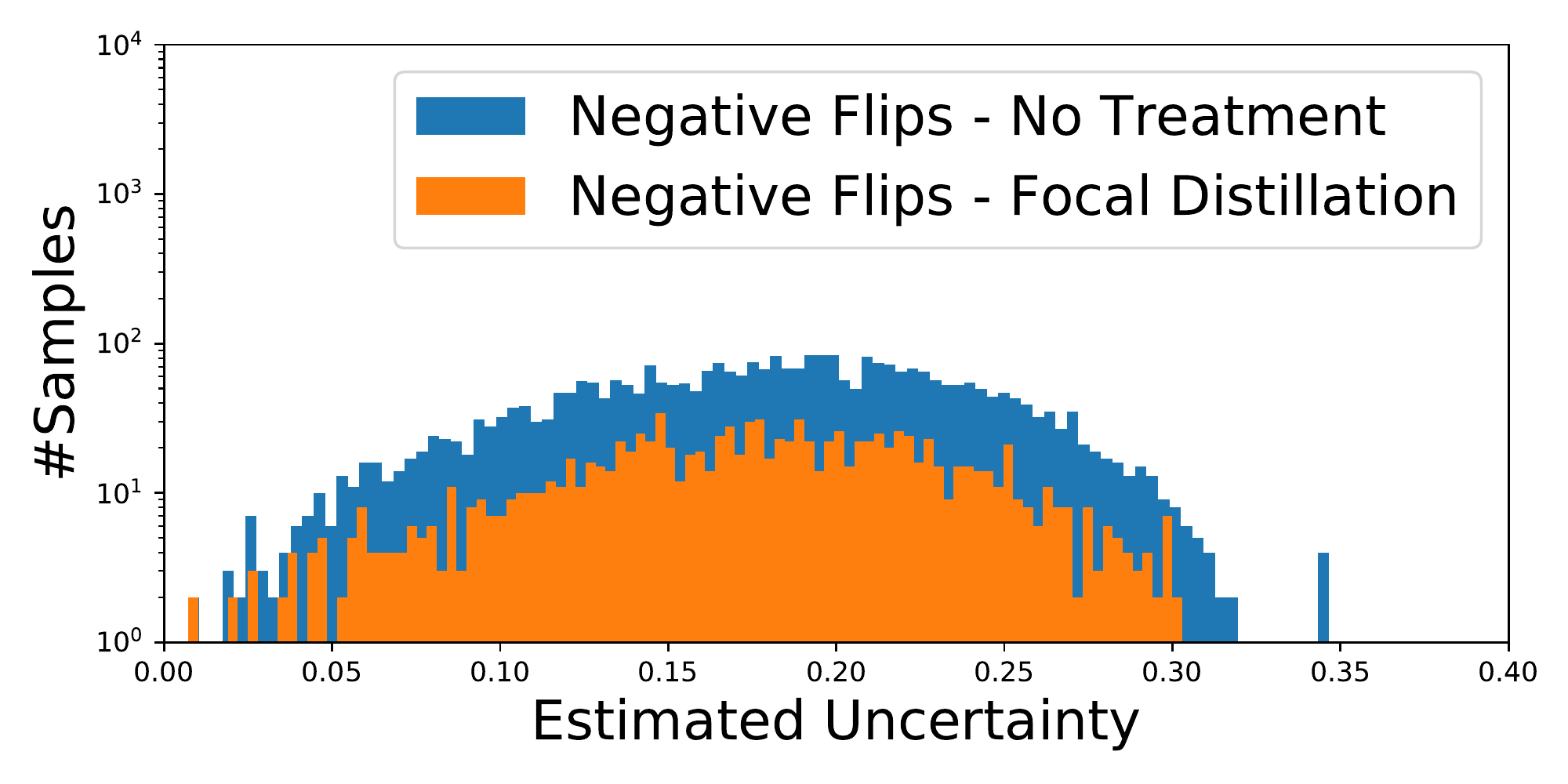}
     \end{subfigure}
    \caption{Distribution of sample uncertainty estimated by~\cite{lakshminarayanan2017uncertainty}, measured on test validation set of ILSVRC12~\cite{ILSVRC15}. Here we use ResNet-18~\cite{he2016resnet} as the model architecture in experiment's. 
    \textbf{Top}: Uncertainty histogram of negative flipped samples and other test samples between two ResNet-18 models trained without PC training. 
    \textbf{Bottom}: Uncertainty histogram of negative flippped sample in two model pairs, the first without PC training. The new model in the second pair is trained with focal distillation based PC training. The $y$-axes are in log-scale. }
    \label{fig:uncertainty}
\end{figure}

\begin{figure}[th]
    \centering
    \includegraphics[width=.80\linewidth]{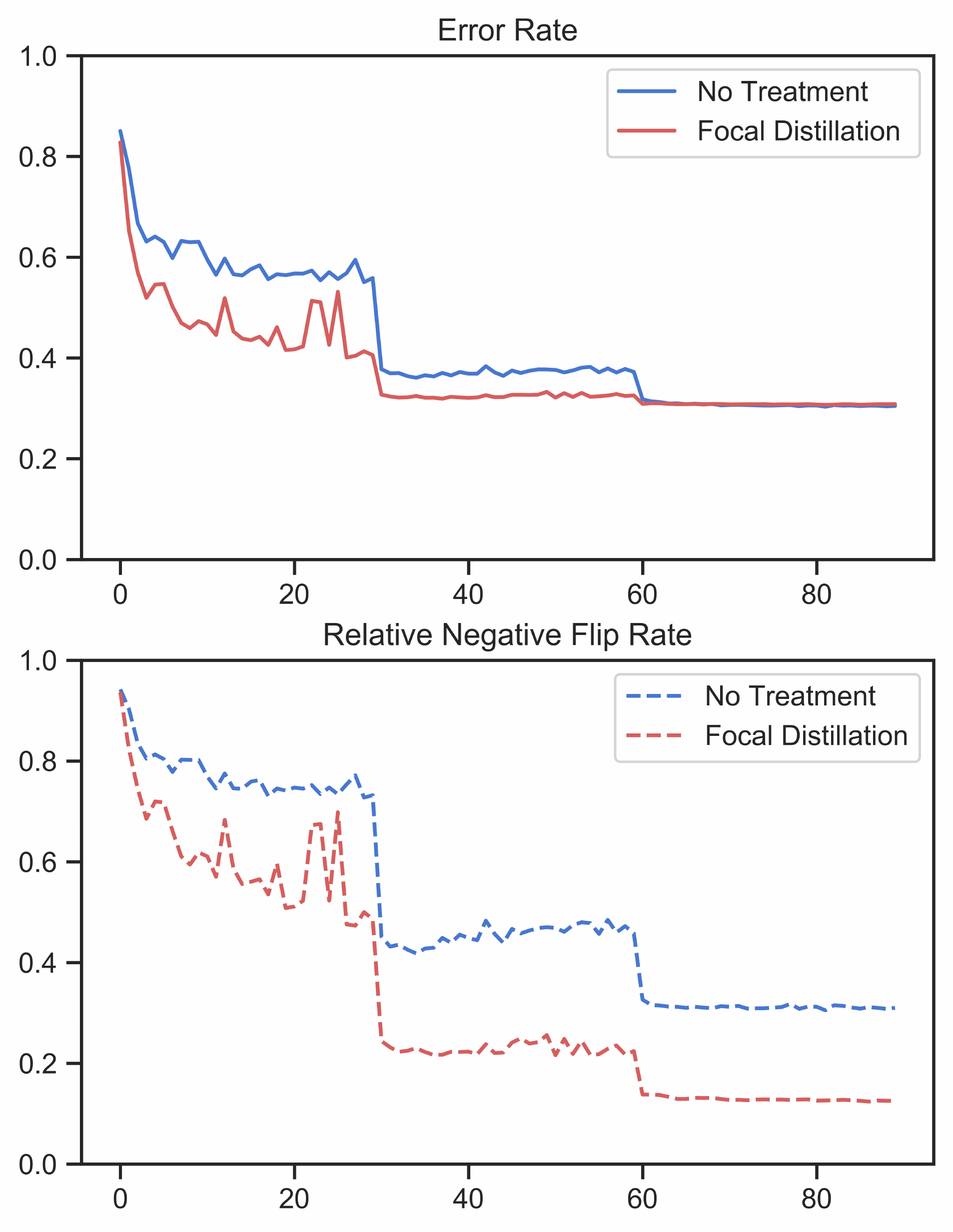}
    \caption{Evolution of error rates and relative NFR during training of the new models on ILSVRC12~\cite{ILSVRC15}. We compare focal distillation-based PC training with no treatment. Generally NFR follows the trend of the error rate during trainig. Focal distillation lead to a gap in relative NFR in early epochs and keeps the gap as training evolves.  }
    \label{fig:nfr_epoch}
\end{figure}

\subsection{Prediction of negative flips}

First we look at the potential relationship between a test sample being a negative flip and its prediction uncertainty. In this experiment we use the uncertainty estimated by a deep ensemble of ResNet-18~\cite{he2016resnet} as described in~\cite{lakshminarayanan2017uncertainty}. We visualize the results in Fig.~\ref{fig:uncertainty}. We observe that test samples that become negative flips have relatively high estimated uncertainty but our measure does not clearly separate them from other samples. We also visualize the histogram of uncertainty estimates before and after we apply focal distillation based PCT, as shown in Fig.~\ref{fig:uncertainty}.  We observe that PCT with focal distillation reduces the NFR with an almost uniform chance for samples with different level of uncertainty. This observation may suggest alternate strategies to designing PC training methods.  

The second study is on the evolution of negative flip rates during the training of the new model. We visualize the relative NFRs and error rates at every epoch when training a new model. We present results for both training without PCT and with focal distillation based PCT (FD-LM) in Fig.~\ref{fig:nfr_epoch}. Our first observation is that the NFRs change in a similar trend as the new models' error rates and reduce as the models are trained longer. By applying PCT during training, we observe that the relative NFR drops faster in early epochs compared with no treatment. As training goes on, the new model with PCT maintains the same gap in NFR until the training terminates.

\section{Discussion}
\label{sec:discussion}

Large-scale DNN-based classifiers are typically only a part of more complex systems that involve additional post-processing. In the simplest cases, the output of the DNN is used to map the data to a metric space, where it is clustered and searched. In more complex cases, classifiers are part of an elaborate system that includes high-level reasoning. In all cases, changing the classifier can break the system, which is why DNN models are seldom updated despite the steady improvements reported in the literature. PC training could enable seamless adoption of improved models, and ensure steady progress and increased accessibility to the state of the art in image classification.
%\textcolor{blue}{PC training aims at minimizing NFR in model updates. It could enable seamless adoption of improved models, and ensure steady progress and increased accessibility to the state of the art in image classification.} 
In this paper, we have merely scratched the surface of PC training, as the methods proposed have obvious limitations: Ensembling is not viable in large-scale systems, even though it achieves the highest NFR reduction. Focal distillation reduces the NFR, but at the price of a slight increase in error rate. Further exploration is needed to identify methods that can achieve paragon performance at the baseline cost of single model distillation.

{\small
\bibliographystyle{ieee_fullname}
\bibliography{main}
}

\clearpage

\begin{appendices}

\section{Two Changes in Model Updates}
Model updates can have two or more changes together. Here we present  results on model updates with both (a) change in model architecture, and (b) increase in training samples. We train a Resnet-18~\cite{he2016resnet} model on 50\% of the classes of the ILSVRC12 training dataset~\cite{ILSVRC15} as the old model. For the new model, we train a Resnet-50~\cite{he2016resnet} on the entire training set. The results are shown in Table~\ref{tab:arch_change_data_growth}. Though the accuracy of the models improve by $ \sim $12\%, 3.56\% of the samples could become negative flips. The different approaches of PC training consistently reduces NFRs.

\begin{table}[h]
\small
\centering
\setlength{\tabcolsep}{1.5mm}{
\begin{tabular}{l||c|c|c|c|c}
\hline
\multirow{2}{*}{Approach}        & \multicolumn{2}{|c|}{Error Rate~(\%)} & \multirow{1}{*}{NFR} & \multirow{1}{*}{Rel. NFR} & \multirow{2}{*}{\#params} \\
\cline{2-3}
& $\phi^{\textnormal{old}}$ & $\phi^{\textnormal{new}}$ &(\%)&(\%) \\
\hline 
No Treatment            &    35.34   &    23.81     &   3.56  & 23.14 & 25M     \\ \hline
Naive                   &    35.69   &    23.30     &   3.12  & 20.82 & 25M     \\ \hline
FD-KD                   &    35.69   &    28.56     &   2.34  & 12.74 & 25M     \\ \hline
FD-LM                   &    35.69   &    27.90     &   2.44  & 13.58 & 25M     \\ \hline\hline
Ensemble                &    31.94   &    21.98     &   2.71  & 18.13 & 100M    \\ \hline
\end{tabular}}
\caption{Change in architecture + increase in \#samples}\label{tab:arch_change_data_growth}
\end{table}

\begin{figure}[t]
    \centering
         \includegraphics[width=0.98\linewidth]{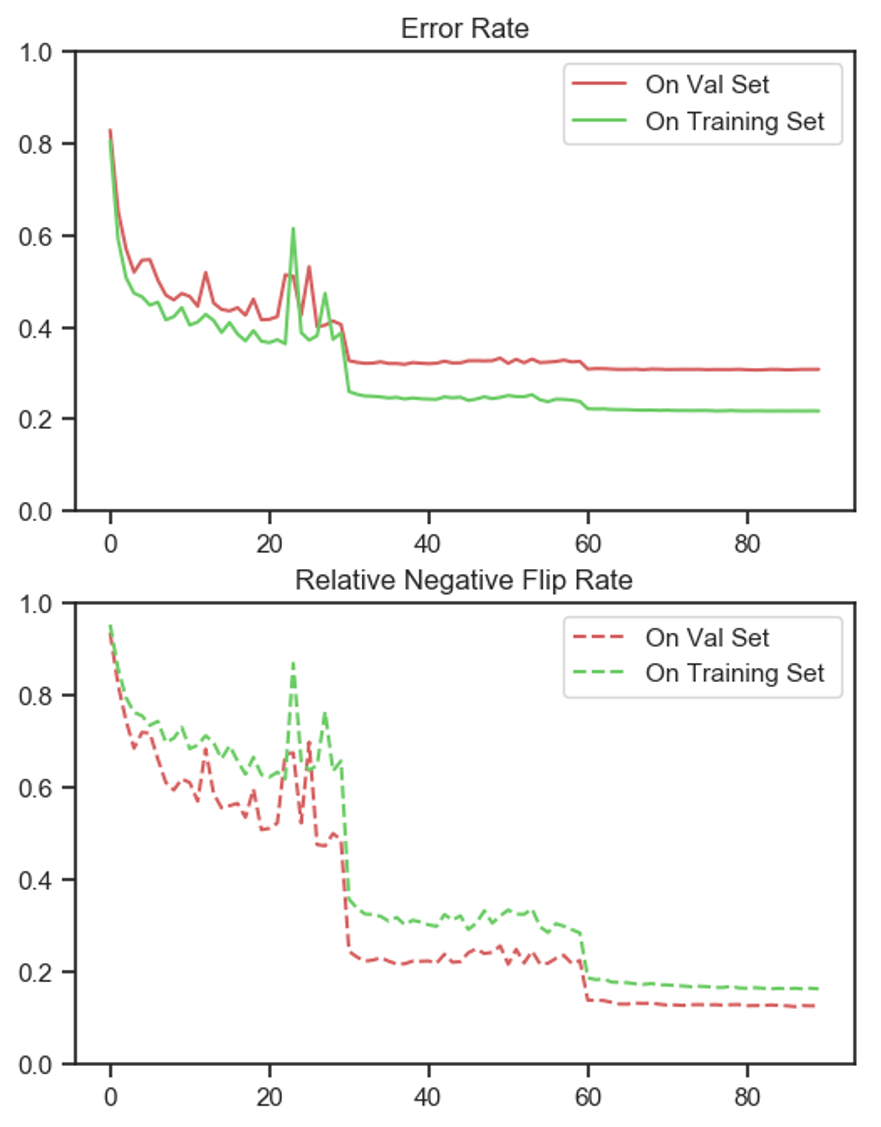}
         \caption{ Relative NFRs and error rates of the new model on both the training and the validataion sets. The new model is trained with focal distillation. We plot the error rates and NFRs after every training epoch to visualize the evolution of the values during training. }
         \label{fig:nfr-on-training-set}
\end{figure}

\begin{table*}[t]
\small
\centering
\setlength{\tabcolsep}{1.5mm}{
\begin{tabular}{l|c||c|c|c|c||c|c|c|c||c}
\hline
\multirow{3}{*}{\bf Approach} & \multirow{3}{*}{\bf Fine-tuning}       & \multicolumn{4}{c||}{\bf Increase in \#samples} & \multicolumn{4}{c||}{\bf Increase in classes} & \multirow{3}{*}{\bf \#params} \\ 
\cline{3-10}
& & \multicolumn{2}{c|}{Error Rate~(\%)} & NFR & Rel. NFR & \multicolumn{2}{c|}{Error Rate~(\%)} & NFR & Rel. NFR & \\
\cline{3-4}\cline{7-8}
& & $\phi^{\textnormal{old}}$ & $\phi^{\textnormal{new}}$ & (\%) & (\%) & $\phi^{\textnormal{old}}$ & $\phi^{\textnormal{new}}$ & (\%) & (\%) &  \\
\hline 
\multirow{2}{*}{No Treatment}            &     \xmark  &    28.58   &    23.65  &  3.98   &  19.47  & 19.30   &  23.65     &  8.05   &  41.90  & 25M     \\ \cline{2-11}
&    \cmark  &    28.58   &    24.92    &     3.28   &    18.42  & 19.30   &  25.31  &  7.96   &  38.97     & 25M     \\ \hline
\multirow{2}{*}{Naive}                   &     \xmark  &    28.45   &    24.46  &  3.27   &  18.68  & 19.28   &  23.74     &  7.70   &  40.20  & 25M     \\ \cline{2-11}
&   \cmark  &    28.45   &    24.50    &     2.78   &    15.85  & 19.28   &  23.69  &  7.14   &  37.33     & 25M     \\ \hline
\multirow{2}{*}{FD-KD}                   &     \xmark  &    28.45   &    25.20  &  2.89   &  15.84  & 19.28   &  24.14     &  7.07   &  36.29  & 25M     \\ \cline{2-11}
&   \cmark  &    28.45    &   25.27    &     2.42   &    13.38  & 19.28   &  24.36  &  6.94   &  35.29     & 25M     \\ \hline
\multirow{2}{*}{FD-LM}                   &     \xmark  &    28.45   &    24.77  &  2.85   &  16.09  & 19.28   &  25.11     &  7.37   &  36.35  & 25M     \\ \cline{2-11}
&    \cmark  &    28.45   &    24.91    &     2.39   &    13.41  & 19.28   &  25.05  &  7.17   &  35.46     & 25M     \\ \hline\hline
\multirow{2}{*}{Ensemble}                &     \xmark  &    26.39   &    22.09  &  2.75   &  16.88  & 17.53   &  21.98     &  6.72   &  37.06  & 100M    \\ \cline{2-11}
&    \cmark  &    26.39   &    22.81    &     2.18   &    12.98  & 17.53   &  23.20  &  6.70   &  35.02     & 100M    \\ \hline
\end{tabular}}
\caption{The special case of finetuning the old model to obtain the new model when there is only data change. This setting does not apply to model updates where the model architecture is changed.}\label{tab:finetuning_data_growth}
\end{table*}

\section{The Special Case of Fine-Tuning}
In model updates only involving data changes but not any architecture change, there is a special case that we can build the new model by finetuning the old model on the new training data. We analyze this special case and compare our PC training approaches in it. We use the same setting as in Sec. 5.3 of the main paper. The results of both scenarios of data changes are summarized in Table~\ref{tab:finetuning_data_growth}. The odd rows denote the normal cases where train the new model from scratch. The even rows denotes the special cases of initializing the new model with the weights of the old model. 
We observe even though the new models with finetuning started with the weights of the old models, they still suffers from negative flips. Applying our PC training approaches help reduce NFR in these special cases similar to that for the normal new models.

\section{Training Set Negative Flips}
By definition, we assess the sample-wise regression problem on the testing samples. It is worth to note that there are also negative flips in the training samples. In Fig.~\ref{fig:nfr-on-training-set} we present the evolution of NFRs and error rates during training of the new model on both the training and the validation sets of ILSVRC12~\cite{ILSVRC15}. The new model uses focal distillation for PC training. Interestingly, the NFR on the training is even higher than that on the validation set.

 \end{appendices}

\end{document}